\documentclass[letterpaper, 10 pt, conference]{ieeeconf}  

\IEEEoverridecommandlockouts                              

\usepackage{graphics} 
\usepackage{epsfig} 
\usepackage{mathptmx} 
\usepackage{times} 
\usepackage{amsmath} 
\usepackage{amssymb}  
\usepackage{booktabs}
\usepackage{svg}
\usepackage{multirow}
\usepackage{tikz}
\usepackage{pgfplots}
\usepackage{color}
\usepackage{cuted}    
\usepackage{capt-of}
\usepackage[colorlinks=true, linkcolor=blue, citecolor=blue, urlcolor=blue, breaklinks=true]{hyperref}
\usepackage{cleveref}
\title{\LARGE \bf
ForwardDLO: Model-Based Bimanual Shape Matching of Unconstrained Deformable Linear Objects}

\author{Tim Missal$^{1}$, Berk Guler$^{1,2}$, Lucas Domingues$^{3}$ \\ Simon Manschitz$^{2}$, Jan Peters$^{1,4-7}$, Paula Dornhofer Paro Costa$^{3,8}$
\thanks{$^1$Technical University of Darmstadt $^2$Honda Research Institute Europe GmbH $^3$School of Electrical and Computer Engineering, Universidade Estadual de Campinas (UNICAMP), Brazil $^4$German Research Center for Artificial Intelligence (DFKI) $^5$hessian.AI $^6$Robotics Institute Germany (RIG) $^7$Centre for Cognitive Science  $^8$Artificial Intelligence Lab, Recod.ai; Corresponding author: \texttt{tim.missal@icloud.com}}}

\begin{document}

\maketitle
\thispagestyle{empty}
\pagestyle{empty}

\begin{strip}
    \vspace{-7.8em}
    \centering
    \includegraphics[width=\textwidth]{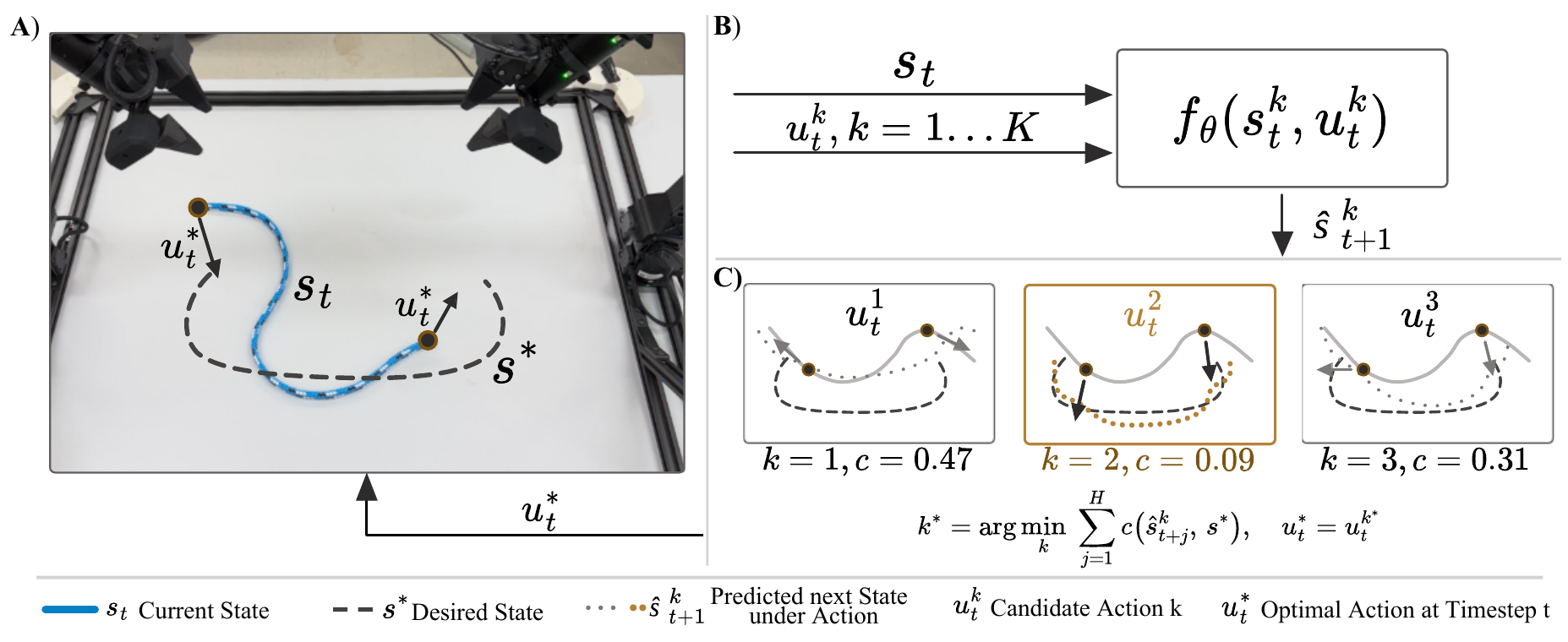}
    \vspace{-2.2em}
    \captionof{figure}{Bimanual shape matching of a deformable linear object (DLO). The DLO is represented by the current shape $s_t$. The goal is to reach the target shape $s^*$.  \textbf{A)}  Two arms grasp it at arbitrary segments and apply a bimanual action $u_t$.  \textbf{B)} A learned forward model $f_\theta$ predicts the next shape $\hat{s}^k_{t+1}$ for each of $K$ sampled candidate actions $u^k_t$. \textbf{C)} Every rollout is scored against the target by the shape matching cost $c$, which can vary depending on the task but is defined as the root mean square error in our experiments (\hyperref[sec:results]{Sec. V}). The lowest-cost candidate (gold) gives $u^*_t$, which is executed on both arms before replanning.}
    \label{fig:hero}
    \vspace{-1.7em}

\end{strip}
\begin{abstract}
Ropes, cables, and other deformable linear objects appear in tasks from untangling to cable routing and suturing, yet controlling their shape remains a challenge in robot manipulation. We study model-based shape control in a general setting: the object lies unfixated on a support surface and two arms may grasp and move it anywhere along its length. Because each arm chooses a grasp point, direction, and magnitude, the joint action space is combinatorially large, and the dynamics model's per-prediction cost bounds how much of it a planner can search. We present ForwardDLO, a recurrent latent dynamics model for this unfixated bimanual setting that predicts per-segment displacements grounded in the observed rope state at every step. Our model reaches accuracy comparable to more expensive baselines while containing no explicit segment-to-segment operations, which makes batched evaluation of candidate actions cheap. On open-loop prediction of real rope motion it reaches the lowest error of the learned models we evaluate, 13\% below the strongest baseline. Within a fixed time budget it scores 8 to 22 times more candidate actions than models of comparable accuracy while matching them in real-world shape matching; and on a simulated routing task at a 30\,Hz control rate, this throughput converts into 98\% task success versus at most 30\% for the baselines at their own budgets. We release the model, code, and a dataset of 2.42 million simulated and 14{,}107 real rope transitions in \href{https://anonymous.4open.science/r/ForwardDLO/}{the project's repository}.
\end{abstract}

\section{INTRODUCTION}
\label{sec:intro}
Deformable Linear Objects (DLOs), such as ropes and cables, are
ubiquitous in everyday life and play a central role in target
application domains for robotics, such as routing a cable through a
harness in industry~\cite{malvidofresnilloDualarmRoboticSystem2025}. Yet manipulating DLOs remains a challenging problem
in robotics due to their highly coupled dynamics and nearly infinite
degrees of freedom.

Many robotic applications involving DLOs require controlling their shape to match a desired one. This shape might be a two-dimensional form or a three-dimensional knot~\cite{freundTWISTEDRLHierarchicalSkilled2026}.
Reaching such a shape from an arbitrary starting configuration is an
instance of the classic control problem: given the current state, how do
we choose the actions that drive the system toward a desired goal?

There are two ways to address this problem. Given a specific goal, one
technique is to define specialized policies that choose actions to reach
that goal \cite{wigginghausLearningSimGroundedPolicies2026}. Another
approach is to plan using forward models. With an accurate model of an
environment, one can evaluate candidate actions and resulting states
under a loss function. Given an accurate enough model, this approach can
in principle solve the control problem for any system so described,
though computing the optimum is generally intractable and practical
methods rely on approximate
solvers~\cite{bertsekasDynamicProgrammingOptimal1995}. Learned models for DLOs have been studied in two complementary settings:
unimanual manipulation of a free
rope~\cite{yanSelfSupervisedLearningState2020,zhangDeformableLinearObject2021,leeSampleEfficientLearningDeformable2022}, and manipulation of a rope
fixated at one or both
ends~\cite{yangLearningPropagateInteraction2021,guLearningGraphDynamics2025,wangOfflineOnlineLearningDeformation2022,yuShapeControlDeformable2022,yuGlobalModelLearning2023,yueLSTMGCNHybridArchitecture2025}.

Tasks such as tying a knot~\cite{gulerAssistDLOAssistiveTeleoperation2026} or closing a wound with sutures motivate the study of settings in which a DLO is not fixed, can be grasped at different points along its length, and is manipulated using two arms. Addressing such settings benefits from models that provide accurate short- and long-term predictions while remaining efficient enough to evaluate the comparatively large action space of bimanual manipulation.

To move model-based DLO manipulation toward these settings, we make
three contributions:

\begin{itemize}
    \item ForwardDLO, a learned dynamics model for DLOs that
    outperforms more expensive models on out-of-distribution open-loop
    prediction and matches them in Model Predictive Control (MPC). It does so while achieving 8--22$\times$ higher throughput than the models closest to it in
    accuracy, enabling the exploration of high-dimensional action
    spaces.
    \item An evaluation of learned dynamics models on an unconstrained
    DLO under bimanual actions. To our knowledge, this is the first
    evaluation of learned dynamics models in this setting.
    \item A dataset of simulated and real rope behavior containing
    2.43\,M transitions (2.42\,M simulated, 14{,}107 real) that can be
    used for the training and evaluation of DLO models under these
    conditions. Our dataset, along with our code and trained weights for
    our model and all baselines, is available in \href{https://anonymous.4open.science/r/ForwardDLO/}{the project's repository}.
\end{itemize}

\section{RELATED WORK}
\label{sec:related}
We summarize prior work on modeling DLO dynamics for predictive control, organized around predictive model class and task assumptions.

\subsection{Predictive Models}
Physics-based simulators predict DLO motion by integrating a mechanical model:
position-based dynamics (PBD)~\cite{mullerPositionBasedDynamics2007} and its
extension XPBD~\cite{macklinXPBDPositionbasedSimulation2016} project constraints directly onto
positions, discrete elastic rods~\cite{bergouDiscreteElasticRods2008} and
Cosserat formulations~\cite{zhaoEfficientStableSimulation2022} resolve bending and twisting explicitly, and
articulated formulations represent the DLO as a serial chain of rigid links.
Their classical drawback is that behavior depends on material parameters that
must be identified. Resolving constraints and contact for every candidate
action is also typically more expensive than a single forward pass of a learned
model. Among data-driven models, \textit{segment-level} approaches predict motion of single points along the DLO: IN-BiLSTM~\cite{yangLearningPropagateInteraction2021} combines an interaction network for pairwise segment effects with a bidirectional LSTM that propagates them along the chain, LSTM-GCN~\cite{yueLSTMGCNHybridArchitecture2025} pairs recurrence with graph convolutions and Wang et al.~\cite{wangOfflineOnlineLearningDeformation2022} train a graph network offline and correct its residual with an online local model. \emph{Jacobian-based} models instead map end-effector velocity to segment velocity through a state-dependent Jacobian; learned globally in the DLO's configuration space~\cite{yuGlobalModelLearning2023}, they are cheap and reach large deformations through many small steps, but they are derived for an elastic DLO held continuously at its ends and therefore do not extend to a rope that is released, settles under friction, or is re-grasped at an interior point. \emph{Image-space} models predict future observations: Zhang et al.~\cite{zhangDeformableLinearObject2021} fit locally linear dynamics in the latent space, and Lee et al.~\cite{leeSampleEfficientLearningDeformable2022} learn the forward model directly in image space from self-supervised real-world data. They operate in the unfixated, arbitrary-grasp setting; their predictions
are made in image space and are therefore tied to the viewpoint and
appearance of the training scenes, whereas we model segment positions,
which are camera- and appearance-agnostic.

A separate line learns dynamics in a latent state space rather than at
the segment level. RopeDreamer~\cite{missalRopeDreamerKinematicRecurrent2026}
adapts the recurrent state-space model of
Hafner et al.~\cite{hafnerLearningLatentDynamics2019} to DLOs, encoding
the observed rope state and the action into a recurrent latent belief
and decoding full future states from it. We adopt the same latent belief
but differ in three respects. First, we use no observation or action encoders: the belief reads
the rope state directly and the action enters the recurrence
as the raw command vector. Second, in place of a single decoder that
predicts the entire rope state at once, a single decoder with shared
weights is applied independently at each segment. Third, the predicted state is
decoded explicitly at every step and fed back as the next input, so
rollouts pass through explicit rope shapes rather than being carried in
latent space, and an observation can replace a prediction at any step.

\begin{figure*}[t]
    \centering
    \includegraphics[width=\textwidth]{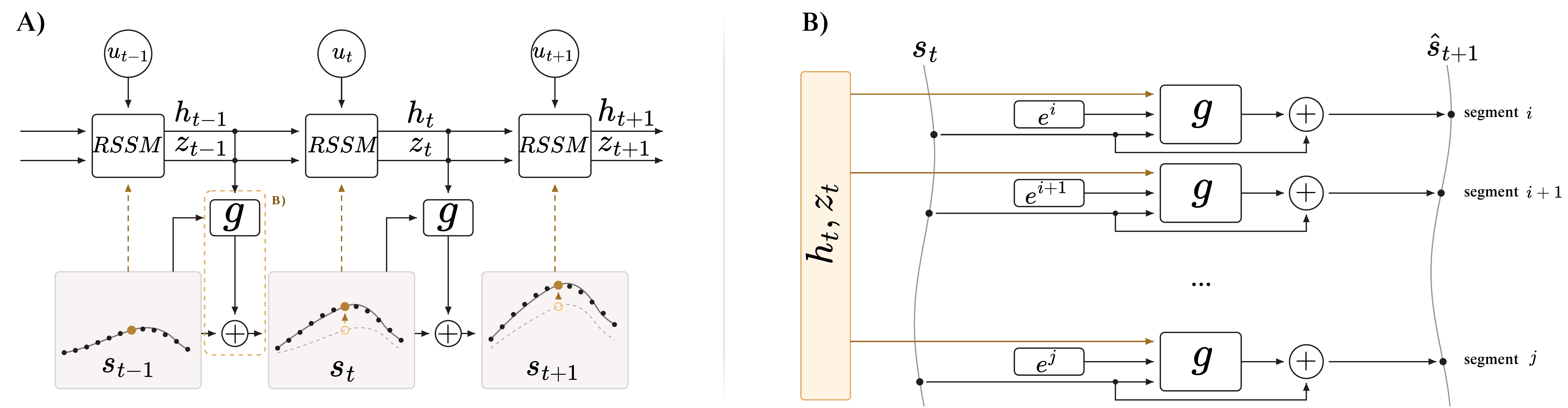}
    \caption{Recurrent latent dynamics model with a shared per-segment decoder. Parts that are unused during inference are not shown. \textbf{(A)}~The model unrolled over three steps. At each step the recurrent state-space model (RSSM) consumes the previous recurrent state $h_t$ and stochastic latent $z_t$ together with the bimanual action $u_t$ and the current (predicted) rope state $s_t$, and emits the next $(h,z)$. The decoder $g$ predicts a per-segment displacement that is added residually to the current segment positions to give $\hat{s}_{t+1}$. During planner rollouts, the model re-grounds on its own prediction rather than on an observation (self-observe, gold). \textbf{(B)}~Detail of the decoder for one step. The state $(h_t,z_t)$ is
    broadcast to every segment, and a single decoder $g$ with shared weights is
    applied independently at each segment $i$, taking that segment's own position
    as the residual base and a learned segment embedding $e^{(i)}$ that identifies it.
    There are no explicit segment-to-segment operations: segments interact only through the
    shared state, so all spatial coupling along the rope is mediated by $(h,z)$.}
    \label{fig:model}
    \vspace{-1.7em}
\end{figure*}

\subsection{Task Assumptions}

Approaches to DLO shape matching differ mainly in three assumptions: whether the DLO is fixated, where along its length it may be grasped, and how many arms act on it. Most work holds one or both ends and manipulates only at those ends~\cite{yangLearningPropagateInteraction2021,guLearningGraphDynamics2025,wangOfflineOnlineLearningDeformation2022,yuShapeControlDeformable2022,yuGlobalModelLearning2023,yueLSTMGCNHybridArchitecture2025}. Holding the ends matches applications where the DLO is attached or being
inserted, and it makes the configuration largely determined by the two
boundary poses; targets that require moving the object as a whole or
acting at interior points lie outside this setting by construction. The complementary line places an unfixated DLO on a plane and grasps it anywhere along its length, it reaches far more target shapes but is unimanual throughout~\cite{yanSelfSupervisedLearningState2020,zhangDeformableLinearObject2021,leeSampleEfficientLearningDeformable2022, missalRopeDreamerKinematicRecurrent2026}. The two settings are disjoint. Shape matching for tasks such as untangling, routing, or suturing requires their union, which we are aiming for in this work.

Bimanual work that learns a general DLO model grasps the two ends, where the two boundary poses again constrain the configuration~\cite{wangOfflineOnlineLearningDeformation2022,yueLSTMGCNHybridArchitecture2025,yuGlobalModelLearning2023}. The models learned in the unfixated, arbitrary-grasp setting are unimanual and quasi-static~\cite{yanSelfSupervisedLearningState2020,zhangDeformableLinearObject2021,leeSampleEfficientLearningDeformable2022}. To our knowledge, no learned dynamics model has been evaluated for bimanual shape matching of an unfixated DLO. In our work, we close this gap by evaluating ForwardDLO, as well as other baselines, in this setting.

\section{PROBLEM FORMULATION}
\label{sec:problem}

We consider the task of modeling the dynamics of a DLO under quasi-static
manipulation by two robotic end-effectors, as shown in
\hyperref[fig:hero]{Fig.~1A}. The DLO rests on a horizontal support
plane under gravity and is assumed to have low bending stiffness, admitting a wide range of shapes. We treat it as a free-moving multi-body system
and represent it by $N$ equidistant segments, each with a position in
$\mathbb{R}^3$.
Actions displace the DLO out of the support plane, so the state is
three-dimensional throughout.

The state of the DLO at timestep $t$ is
$s_t = [\,s^{(0)}_t, \dots, s^{(N-1)}_t\,] \in \mathbb{R}^{3 \times N}$,
where $s^{(i)}_t \in \mathbb{R}^3$ is the position of segment $i$.

The robotic interaction is a bimanual pick-and-place action
$u_t = (u^{L}_t,\, u^{R}_t)$ with one component per gripper,
$u^A_t = (a^A_t,\, i^A_t,\, \delta^A_t)$ for $A \in \{\mathrm{Left}, \mathrm{Right}\}$.
Each active gripper performs a grasp, a vertical lift, a translation in the
$XY$-plane at the lift height, and a descent back to the support surface,
after which the DLO settles to rest. The indicator $a^A_t \in \{0,1\}$
states whether gripper $A$ acts at time $t$; if it does not, $i^A_t$ and
$\delta^A_t$ are ignored. The index $i^A_t \in \{0, \dots, N-1\}$ is the
segment grasped by $A$, and
$\delta^A_t = [\delta^A_x,\, \delta^A_y]^\top \in \mathbb{R}^2$
is the in-plane displacement applied to that segment during the translation
phase. Because the segment is lifted to a fixed height before it is
translated, the rope moves in $\mathbb{R}^3$ and can cross over itself,
even though $\delta^A_t$ is planar.

We seek a model $f_\theta$ that predicts the DLO state resulting from a given
action and the history of past states and actions:

\vspace{-1em}
\begin{equation}
    \hat{s}_{t+1} = f_\theta(s_{0:t},\, u_{0:t}).
\end{equation}

\section{METHOD}
\label{sec:method}

We contribute a large-scale dataset of simulated and real-world DLO transitions, enabling controlled training and evaluation of dynamics models and facilitating reproducible comparisons between simulation and real-world performance.  We also introduce ForwardDLO, a learned dynamics model for DLOs that predicts how a DLO behaves under actions using a latent state and a shared per-segment decoder.

\subsection{Dataset}
\label{sec:dataset}

We collect our dataset in simulation using the Newton Engine's~\cite{thenewtoncontributorsNewtonGPUacceleratedPhysics2025} Vertex Block Descent (VBD) \cite{chenVertexBlockDescent2024} simulator. The dataset is collected by randomly placing a rope on a flat ground plane and recording random pick, lift, move, and place actions applied to it. It is structured into 43{,}200 episodes of 56 transitions each for a total of 2.42\,M transitions under a single solver and physics configuration depicted in \hyperref[tab:sim-params]{Table I}. We deliberately train on a single physics configuration and solver, in order to test how well the resulting models transfer to out-of-distribution data such as real-world DLOs of different stiffness.

\begin{table}[h]
\vspace{-0.5em}
\centering
\caption{Simulation parameters}
\label{tab:sim-params}
\small
\vspace{-1em}
\begin{tabular}{@{}ll@{}}
\toprule
Solver           & Newton VBD cable \\
Segments         & $70$ capsules \\
Segment length   & $0.01$\,m \\
Segment radius   & $0.005$\,m \\
Rope length      & $0.70$\,m \\
Stiffness        & $0.005$ \\
Friction    & $0.8$ \\
\bottomrule
\vspace{-2em}
\end{tabular}
\end{table}

Actions are sampled in a 40\% left-hand, 40\% right-hand, 20\% bimanual split. For each arm, a random segment is selected, followed by a randomly sampled direction and displacement length. Grasp indices are drawn uniformly from the $N$ segments. Directions are sampled uniformly from $[0,2\pi)$, and displacement magnitudes are sampled uniformly from $\mathcal{U}(5,30)$ mm. Bimanual grasps are additionally constrained to lie at least ten segments apart to account for manipulator size.

We also collect and publish real-world data using a bimanual ALOHA robot~\cite{zhaoLearningFineGrainedBimanual2023}. The setup is shown in \hyperref[fig:hero]{Fig. 1A}. Data is collected on two different ropes of varying thickness and stiffness. We use a perception pipeline based on SAM3~\cite{carionSAM3Segment2026} for segmentation; from the depth information of the segmented area we interpolate 70 points along the rope. Perception repeatability, measured over ${\sim}240$ pairs of observations of a physically motionless rope, has a median of 3.5\,mm. The dataset consists of two parts. The first contains random exploratory motions totalling 3{,}439 transitions (2{,}024 transitions on the blue rope, 1{,}415 on the white rope). The second and larger part of the dataset is recorded during the closed-loop MPC execution of \hyperref[sec:results]{Sec. V}, where every transition stores the observed state, the executed bimanual action, and the resulting state: 308 runs totalling 10{,}668 transitions (4{,}655 blue, 6{,}013 white). This share is therefore biased toward the goal shapes used in \hyperref[sec:results]{Sec.~\ref{sec:results}} (S, U and J). Individual transitions remain a usable training or fine-tuning signal if shuffled accordingly; multi-step training on this data should be treated with caution, since consecutive transitions within a run are strongly correlated and the distribution of shapes is not uniform. In total, the dataset comprises 14{,}107 real-world transitions.

\subsection{Model}
\label{sec:model}

Predictive models for DLOs must capture how displacing a single segment
affects the object as a whole. Existing models compute these interactions
explicitly, through pairwise neighbor
effects~\cite{yangLearningPropagateInteraction2021} or
attention~\cite{guLearningGraphDynamics2025}, which
faithfully represents local coupling but repeats segment-to-segment
operations at every prediction step. We investigate whether, for an
unconstrained DLO on a support plane, this interaction can instead be
routed through a single recurrent belief, trading explicit message
passing for prediction throughput. The resulting model, ForwardDLO,
contains no explicit segment-to-segment operations, i.e., operations in which the effect that one rope segment being moved has on other rope segments is calculated explicitly.

\paragraph{Model Overview}
One prediction step maps the current rope state $s_t$, the current
belief, and a bimanual action $u_t$ to the next state $\hat{s}_{t+1}$ in
three stages (\hyperref[fig:model]{Fig.~2A}): the current state is
folded into the belief $(h_t, z_t)$ (\emph{belief update}), the belief is advanced
one step under the action $u_t$ (\emph{one-step prediction}), and the advanced
belief $(h_{t+1}, \tilde z_{t+1})$ is decoded by $g$ into one displacement per segment, which is added to the
current segment positions (\emph{per-segment readout}). Two properties
distinguish this design. First, the model never decodes the rope in one
piece: a single decoder with shared weights is evaluated once per segment,
so the per-step cost is $N$ independent evaluations of one function.
Second, the predicted state is decoded explicitly at every step and
re-enters the computation as the next step's decoder input, so rollouts
never leave the state space: errors surface as explicit rope shapes
rather than accumulating in latent space, and an observed state can
replace the prediction at any step.

\paragraph{Belief Update}
The belief is a pair $(h_t, z_t)$, following the recurrent state-space model (RSSM) of Hafner et al.~\cite{hafnerLearningLatentDynamics2019}:
$h_t$ is a deterministic recurrent state and $z_t$ a stochastic latent.
A Gated Recurrent Unit (GRU)~\cite{choLearningPhraseRepresentations2014} advances the deterministic state from the previous latent and the
previously executed action,

\vspace{-1em}

\begin{equation}
  h_t = \mathrm{GRU}\big(h_{t-1},\, [z_{t-1},\, u_{t-1}]\big),
\end{equation}
so, $h_t$ summarizes the interaction history before the current state is
seen. The rope state enters through the latent: a posterior

\vspace{-2em}
\begin{equation}
  q(z_t \mid h_t, s_t) = \mathcal{N}\!\big(\mu_q(h_t, s_t),\,
  \sigma_q(h_t, s_t)\big)
\end{equation}
reads the rope state directly and grounds the belief in the
available state, which is an observation when one exists and the model's
own prediction during rollouts. We abstain from the observation and
action encoders common in latent dynamics models, as ablations
(\hyperref[sec:ablations]{Sec.~V-D}) show them to be redundant here.
$z_t$ is a sample from $q$ during training and its mean at inference.

\paragraph{One-Step Prediction}
To predict the effect of an action $u_t$ before the next state exists,
the same cell advances the belief once more,

\vspace{-1em}
\begin{equation}
  h_{t+1} = \mathrm{GRU}\big(h_t,\, [z_t,\, u_t]\big),
\end{equation}

and, since there is no state to ground on at $t{+}1$ yet, the latent is
drawn from a prior conditioned on the recurrent state alone,

\vspace{-1.7em}
\begin{equation}
  \tilde z_{t+1} \sim p(z_{t+1} \mid h_{t+1}) =
  \mathcal{N}\!\big(\mu_p(h_{t+1}),\, \sigma_p(h_{t+1})\big).
\end{equation}
The prior thus plays for imagination the role the posterior plays for
grounding, and the KL term~\cite{kingmaAutoEncodingVariationalBayes2022} of the training objective keeps the two
consistent, so imagined and grounded beliefs remain interchangeable. The
advanced pair $(h_{t+1}, \tilde z_{t+1})$ is decoded into the next state
by the readout.

\paragraph{Per-Segment Readout}
Rather than decoding
all $3N$ coordinates with one output layer, a single MLP $g$ with shared
weights is evaluated at every segment independently and predicts only the
 displacement of that segment, which is added to its current position,

\vspace{-0.7em}
\begin{equation}
\label{eq:readout}
  \hat{s}_{t+1}^{(i)} = s_t^{(i)} +
  g\big(h_{t+1},\, \tilde z_{t+1},\, e^{(i)},\, s_t^{(i)},\,
  \phi_t^{(i)}\big).
\end{equation}

The belief $(h_{t+1}, \tilde z_{t+1})$ is broadcast identically to every
segment. Since the
shared decoder cannot distinguish the segments by itself, each segment index
$i$ is assigned a learned embedding vector $e^{(i)}$ from a table of $N$
vectors trained with the network, analogous to the learned positional
embeddings of sequence models~\cite{gehringConvolutionalSequenceSequence2017}. 
It allows the shared function to specialize its prediction to the segment
it is applied to; an end segment, for example, responds differently than a
middle segment under the same belief. \hyperref[fig:model]{Fig.~2B} shows the readout. 

The segment's own position
$s_t^{(i)} \in \mathbb{R}^3$ serves as the residual base. 
$\phi_t^{(i)}$ presents the action from the perspective of segment $i$: for
each hand $A$ it contains the
commanded target position of the grasped segment, the signed index distance
$(i - i^A_t)/N$ to the grasped segment, a proximity weight
$\exp(-|i - i^A_t|/\tau)$, and the activity indicator $a^A_t$, with the
entire block zeroed for an inactive hand. The proximity weight makes the
decay of an action's influence along the chain, which
interaction-propagation models compute
explicitly~\cite{yangLearningPropagateInteraction2021}, available to
each segment as a precomputed input; the decay constant $\tau$, in units of
segment indices, sets its length scale.

\begin{figure*}[]
    \centering
    \includegraphics[width=\textwidth]{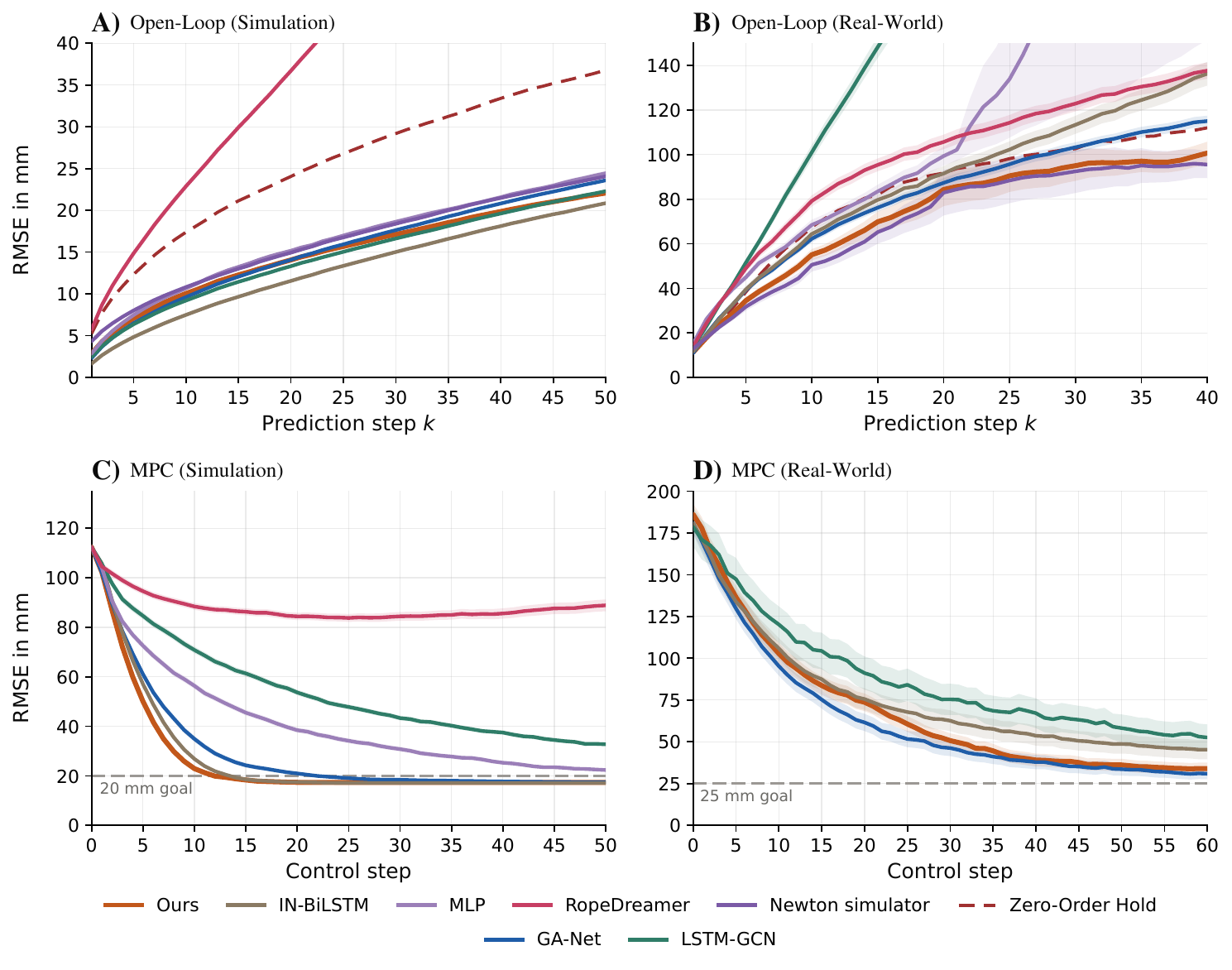}
    \vspace{-2.6em}
    \caption{Open-loop prediction (\textbf{A}, \textbf{B}) and closed-loop MPC
    (\textbf{C}, \textbf{D}) in simulation and in the real world. Curves are means.  \textbf{A)}
    4{,}320 held-out episodes, 50-step rollouts. \textbf{B)} 110 real episodes with at least 40 recorded steps, both
    ropes; bands are standard error over episodes. \textbf{C)} 600 runs
    per model over the shapes S, U and J, stopped at the 20\,mm goal and held.
    \textbf{D)} Real robot, both ropes pooled, 60/56/56/17 runs for Ours/GA-Net/IN-BiLSTM/LSTM-GCN, means adjusted
    for the randomly drawn starting shape by analysis of covariance.}
    \label{fig:mpcopenloop}
    \vspace{-1.7em}
\end{figure*}

\begin{figure}[]
    \centering
    \includegraphics[width=\columnwidth]{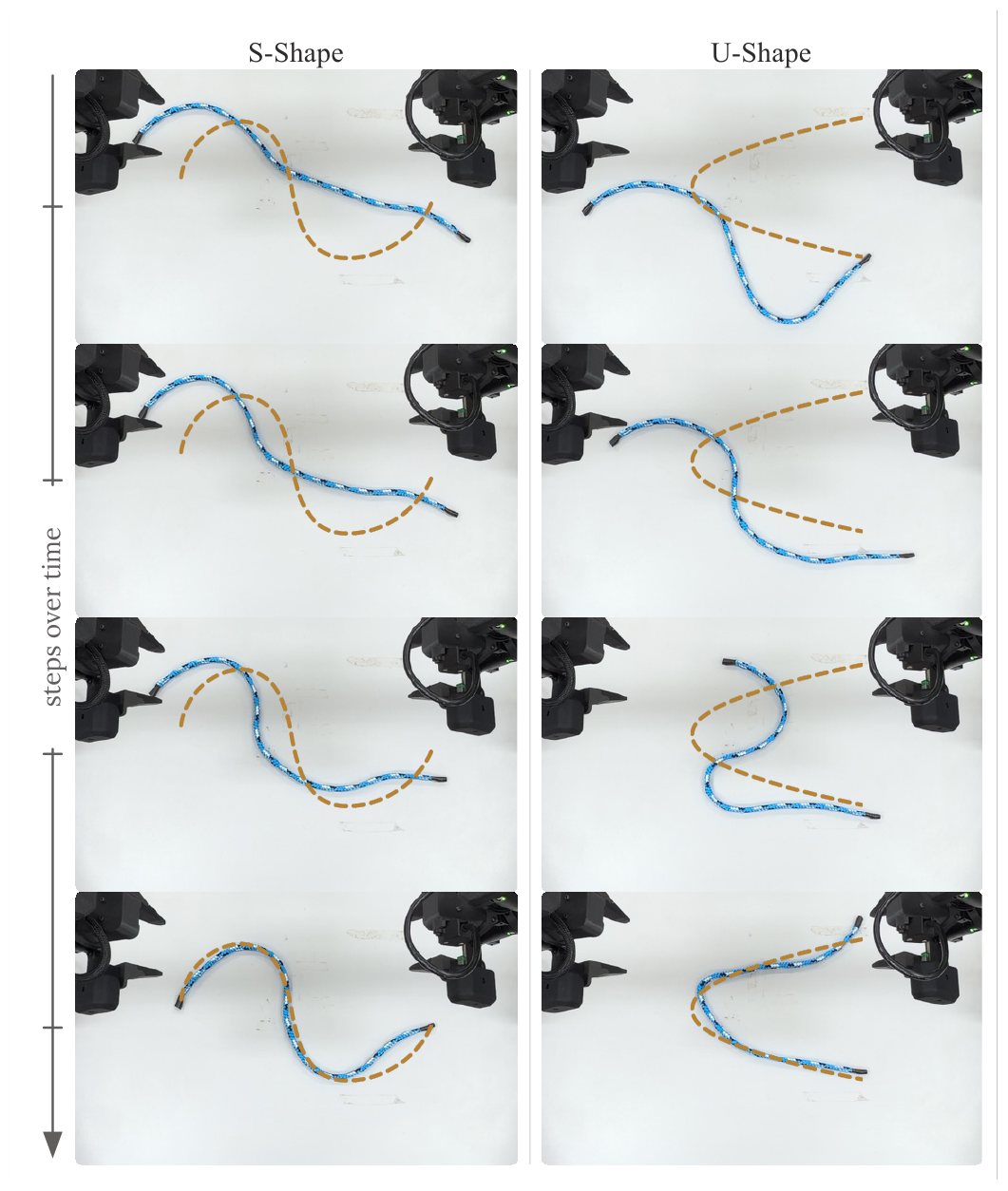}
    \vspace{-1.7em}

    \caption{Real-World Model Predictive Control Rollout of our Model to achieve an S and a U shape.}
    \label{fig:rlmpc}
    \vspace{-1em}
\end{figure}

\paragraph{Multi-Step Rollout}
For horizons beyond one step, the predicted state $\hat{s}_{t+1}$ is fed
back as the next state input (gold arrow in
\hyperref[fig:model]{Fig.~2A}): the posterior grounds the belief on it
as it would on an observation, the cell advances under $u_{t+1}$, and
the readout produces $\hat{s}_{t+2}$. Each step therefore uses both
distributions exactly once.

\paragraph{Training Objective}
The model is trained by filtering each training episode step by step
with the ground-truth states and minimizing, at every timestep,

\vspace{-1.5em}
\begin{align}
  \mathcal{L} = \;& \underbrace{\lVert \hat{s}_t^{\mathrm{rec}} -
  s_t \rVert^2}_{\text{reconstruction}}
  + \underbrace{\lVert \hat{\Delta} s_t - \Delta s_t
  \rVert^2}_{\text{one-step prediction}} \notag \\
  & + \beta\, \underbrace{D_{\mathrm{KL}}\big(q(z_t \mid h_t, s_t)
  \,\Vert\, p(z_t \mid h_t)\big)}_{\text{latent regularizer}}.
\end{align}
The reconstruction head $\hat{s}_t^{\mathrm{rec}} =
d_{\mathrm{rec}}(h_t, z_t)$ requires the grounded belief to explain the
state it has just filtered; it exists only during training and is
discarded at inference. The prediction term supervises the quantity used
at inference: $\Delta s_t = s_{t+1} - s_t$ is the true per-segment
displacement and $\hat{\Delta} s_t$ the readout's prediction of it,
computed from the advanced belief as in \hyperref[eq:readout]{Eq.~(6)}. The KL term is the
usual variational
regularizer~\cite{kingmaAutoEncodingVariationalBayes2022}, with $\beta$
annealed linearly over the first epochs.

\section{RESULTS}
\label{sec:results}

We evaluate our model along three axes: open-loop prediction of rope
deformation, closed-loop shape control with MPC, and planning throughput,
each in simulation and on the real robot. Since we train on a single
simulated physics configuration, we treat simulated performance as a proxy
for in-distribution performance, i.e., settings where the physics
parameters are known or well estimated (e.g., by an external
estimator), or where fine-tuning
is possible. Real-world data is in this sense inherently
out-of-distribution, and evaluation on it measures transfer.

We compare against GA-Net~\cite{guLearningGraphDynamics2025},
IN-BiLSTM~\cite{yangLearningPropagateInteraction2021},
LSTM-GCN~\cite{yueLSTMGCNHybridArchitecture2025}, RopeDreamer~\cite{missalRopeDreamerKinematicRecurrent2026} and a 3-layer MLP, trained to convergence at
the model sizes and learning rates of their original publications.

We quantify accuracy by the root mean square error (RMSE) between predicted and
ground-truth segment positions, averaged over segments and coordinates.

\begin{table}[h]
\centering
\caption{Model and training parameters}
\label{tab:model-training-params}
\small
\vspace{-1em}
\begin{tabular}{@{}ll@{}}
\toprule
Deterministic state $h_t$   & $96$ \\
Stochastic latent $z_t$     & $32$ \\
Prior / posterior head width & $160$ \\
Segment embedding $e^{(i)}$    & $160$ \\
Per-segment head $g$           & $3$ layers, width $160$ \\
Proximity constant $\tau$   & $5$ \\
Parameters, total           & $369{,}973$ \\
\quad active at inference   & $289{,}763$ \\
\quad recon.\ decoder (train only) & $80{,}210$ \\
\bottomrule
\end{tabular}
\vspace{-2em}
\end{table}

\hyperref[tab:model-training-params]{Table II} lists all model parameters. We further use the Adam optimizer with a learning rate of $5\times10^{-4}$, batch size $32$ and an epoch budget of $200$. $\beta$ is increased from $0 \rightarrow 1$ over $5$ epochs.

\subsection{Open-Loop Prediction}
\label{sec:openloop}

We evaluate the accuracy of long autoregressive rollouts in simulation and
in the real world (\hyperref[fig:mpcopenloop]{Fig. 3A/B}), where the real-world dataset represents the
subset of our dataset that only includes episodes of length 40 and above. Each model predicts the next
state from its own previous prediction, and the commanded drag vectors are
re-applied to the rollout's grasped segment, so no ground-truth information
enters after the initial state. As a reference we report a zero-order
hold, which predicts that the rope simply does not move.

In simulation all learned models except RopeDreamer remain well below the zero-order
hold over the full horizon and end within 3\,mm of each other; IN-BiLSTM
is the most accurate. The
differences are statistically significant over the 4{,}320 test episodes
but amount to one to two millimeters. The simulator itself, replayed under
the identical protocol, trails the models trained on its own output. This could be attributed to the solver being re-initialized
from segment positions alone and not being able to recover its internal solver state,
whereas the learned models are trained to predict from exactly this
information.

In the real world all errors multiply. The real ropes
are stiffer than the simulated one, so the dynamics are out of
distribution and the same action displaces more of the rope, which is also visible in the growth of the zero-order-hold reference. At $t{=}1$, ForwardDLO, GA-Net and IN-BiLSTM perform best of all learned models, but differences between them are not significant. After a couple of steps, our model and the
simulator remain clearly below other baselines, with GA-Net notably also staying below the zero-order hold reference for the whole episode. Although IN-BiLSTM, LSTM-GCN and the MLP showed promising performance on simulated data, their errors grow past the zero-order hold. Interestingly, RopeDreamer does not diverge on real data, while still staying above the zero-order-hold reference.

\begin{figure*}[]
    \centering
    \includegraphics[width=\textwidth]{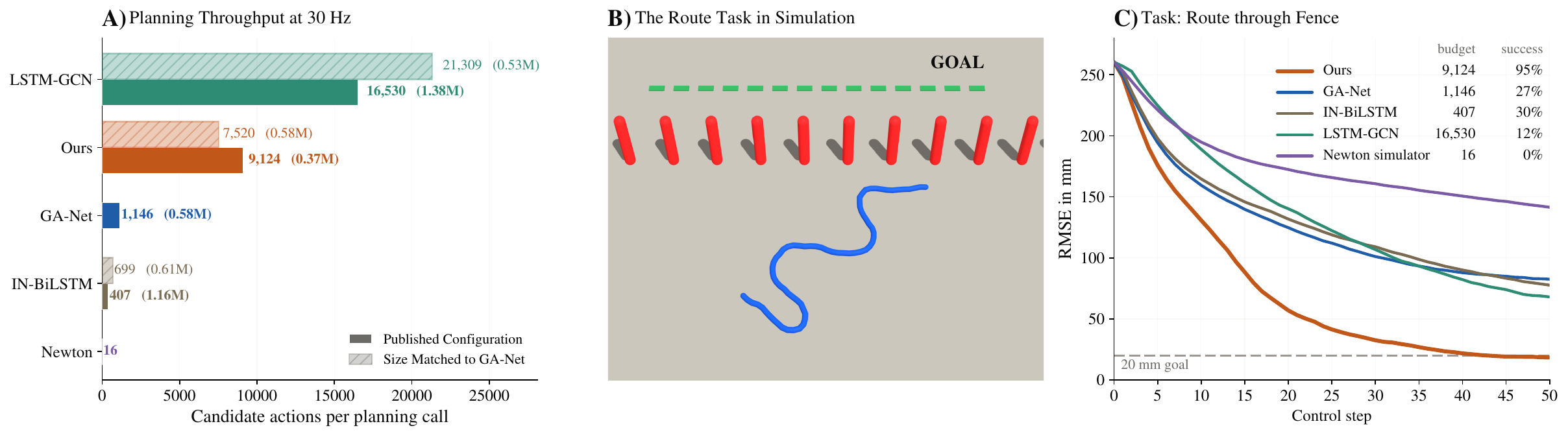}
    \vspace{-2.3em}
    \caption{Planning throughput and its task consequences.
    \textbf{(A)}~Candidate actions each dynamics model can evaluate within
    one 30\,Hz control period, at the published configuration and at a size matched to GA-Net. Timings were measured on an RTX~4060~Ti. Absolute rates are hardware-dependent, but the relative magnitudes were
    roughly constant across the five GPUs we tested on (NVIDIA L40S, Quadro RTX~8000,
    Quadro RTX~5000, RTX~4060~Ti, Quadro P5000). All timings use the same harness for every model. We batch 4{,}225 actions over 30 timed repetitions. Increasing batch size favors ForwardDLO and LSTM-GCN more than other baselines.
    \textbf{(B)}~The route task in simulation. The rope (blue) starts in front
    of a fence of posts (red) and has to be moved onto the goal line (green). The posts are solid obstacles in the
    simulator, but no model is told about them. Each model only predicts how
    the rope moves, so the same checkpoints as for previous tests are used. Avoiding the posts is left to the planner, which discards
    any predicted rope shape that would run into a post. 
    \textbf{(C)}~Closed-loop MPC on the route task (60\,mm post clearance):
    mean RMSE to the goal shape over control steps, each model using the same planner at its own 30\,Hz budget from~A. The legend gives each model's
    budget and success rate (goal reached, $\leq 20$\,mm).}
    \label{fig:throughput}
    \vspace{-1.7em}

\end{figure*}

\subsection{Bimanual Shape Matching}
\label{sec:mpc}

We evaluate closed-loop control on three target shapes, S, U and J, in
simulation and in the real world (\hyperref[fig:mpcopenloop]{Fig. 3C/D}). Every model serves
as the dynamics model under the same simple planner. The planner first samples $10^4$ random candidate actions (a grasp segment and a drag
vector per hand), then plans through the models using a horizon of 1 as early tests showed planar shape matching does not seem to benefit
from long-term planning. After, each predicted state is evaluated under the cost function $c = \mathrm{RMSE}$ and the best one is executed. Closed-loop experiments
    report distributions over 600 simulated environments and 56--60 physical
    runs per model; LSTM-GCN was stopped after 17 runs, at which point its
deficit was already statistically significant (Fisher exact, $p<0.05$). We test significance with paired Wilcoxon tests in simulation,
where all models share environments and goals, and Mann--Whitney U tests
in the real world, Holm-corrected. A real-world run is halted once its
best error has not improved for 15 consecutive control steps;
\hyperref[fig:rlmpc]{Fig. 4} shows complete rollouts of our model on the S and U
targets.

In simulation our model converges fastest, reaching the 20\,mm goal in a
median of 10 control steps against 13 for GA-Net and IN-BiLSTM
($p{<}10^{-38}$), and all three reach it in every environment. After
convergence the three settle within a millimetre of each other, so the
difference lies in convergence speed rather than attainable accuracy.
LSTM-GCN reaches the goal in 26\% of environments within 50 steps.
The simulator itself, used as the planner's dynamics model under the
identical candidate set, matches the learned models' convergence and
plateau while requiring orders of magnitude more planning time per step
(\hyperref[sec:planning-time]{Sec. V-C}).

In the real world the three stronger models are indistinguishable. Our model reaches the goal in 80\% of 60 runs, GA-Net in 86\% of 56, and IN-BiLSTM in 64\% of 56, with median final errors of 24.4, 23.7 and 25.1\,mm; no comparison among them is significant, at any control step of \hyperref[fig:mpcopenloop]{Fig.~3D}, in per-run final error (Mann--Whitney, Holm-corrected $p\approx0.61$ against both) or in success rate (Fisher exact, $p=0.63$ and $p=0.07$). LSTM-GCN reaches its goal in 5 of 17 runs and is significantly worse than all three (Fisher exact, $p<0.05$). We do
not run the simulator as a dynamics model in the real world due to its long planning time, but we expect it to perform on a similar level as the best learned models.

\subsection{Planning under Time Constraints}
\label{sec:planning-time}

The accuracy differences among the three strongest models are small; their difference in computational cost is not. The
comparison of \hyperref[sec:mpc]{Sec.~V-B} holds the candidate set fixed,
which isolates prediction quality but charges nothing for evaluation time.
A deployed planner faces the opposite situation, where the control rate
fixes the time per planning call and the model determines how many actions
fit into it. \hyperref[fig:throughput]{Fig.~\ref{fig:throughput}A} reports
this throughput at 30\,Hz: among the most accurate models on real rope
data, ours evaluates an order of magnitude more actions in the same
window. This allows our model to explore an action space with significantly tighter coverage in the same timeframe.

Whether this matters depends on the task. We construct a task where we expect candidate
density to be the binding resource: guiding a rope through gaps in a fence of posts to a goal line (\hyperref[fig:throughput]{Fig. 5B}), which we treat as a simple proxy for realistic tasks in industry such as wiring a harness~\cite{malvidofresnilloDualarmRoboticSystem2025}. We use the same random-shooting
planner and every model gets its 30\,Hz budget. Here, throughput decides the outcome: our model succeeds in 98\% of 200 runs, against 29.5\% for GA-Net, 30\% for IN-BiLSTM, 12.5\% for LSTM-GCN and ${\sim}0\%$ for the simulator (\hyperref[fig:throughput]{Fig.~\ref{fig:throughput}B,\,C}). A
matched-count control at 9{,}124 candidates attributes the gaps: IN-BiLSTM
and the Newton simulator recover to 90\% and 96\%, so their deficit is almost entirely throughput;
GA-Net improves to 54\%, leaving a residual accuracy gap; LSTM-GCN is unchanged. We further find that changing random shooting to the cross-entropy method (CEM) \cite{rubinsteinCrossEntropyMethodCombinatorial} with 5 iterations does not change this ordering, with LSTM-GCN improving slightly and IN-BiLSTM, GA-Net and ours degrading slightly. An accurate model is thus
necessary but not sufficient for specific tasks: where a planner must distinguish among many
similar actions within one control step, the throughput gap between
equally accurate models decides which of them remain usable at all.

\subsection{Ablations}
\label{sec:ablations}

We perform ablation studies on ForwardDLO to assess whether parts of our model
are redundant and how much each component contributes to its performance. All
ablations are trained on the same dataset as the base model and evaluated on
the same 110 real-world episodes with at least 40 steps in open-loop prediction. For reference,
the base model reaches an RMSE of 100.3\,mm at $t{=}40$. We test each
ablation against the base model with a paired per-episode Wilcoxon signed-rank
test at $t{=}40$; all differences reported as significant remain so after Holm
correction across the comparisons of this section ($p<0.01$).

Reducing the depth of the per-segment decoder degrades accuracy: at $t{=}40$
the error rises to 168.4\,mm with a single layer ($p<10^{-25}$) and
107.3\,mm with two ($p=10^{-3}$). A fourth layer
brings no significant change (99.2\,mm, $p=0.83$).
Replacing the RSSM with a GRU of equal state size increases the real-rope
error to 116.3\,mm ($p<10^{-4}$), while test error in simulation is unaffected (22.1 vs.\
22.3\,mm at $t{=}50$). This suggests the RSSM is redundant on well-estimated
physics while serving as a regulator on out-of-distribution data.

We further study which differences from RopeDreamer, our closest architectural
baseline, contribute to ForwardDLO's performance. Adding explicit state and
action encoders as in RopeDreamer does not change performance (101.1\,mm,
$p=0.70$). Decoding all segment positions with a single MLP instead of the
shared per-segment decoder yields comparable short-horizon accuracy (52.8
vs.\ 54.4\,mm at $t{=}10$, $p=0.75$) but larger long-horizon error
(140.2 vs.\ 100.3\,mm at $t{=}40$, $p<10^{-12}$). Removing the
re-grounding on the model's own prediction leaves short horizons unaffected
and accumulates a uniform deficit beyond $k\approx20$, ending at 109.4\,mm
($p<10^{-4}$).

\section{CONCLUSION AND FUTURE WORK}
\label{sec:conclusion}

We presented a recurrent state-space model for bimanual rope manipulation.
Trained purely in simulation, the model matches the prediction accuracy
of the baselines in simulation, and on real-world data it outperforms
learned baselines and matches the simulator itself. In closed-loop shape
matching on a physical robot it is statistically indistinguishable from
the strongest baselines, while evaluating 8 to 22 times more candidate
actions per planning call than the baselines of comparable accuracy, and
over 500 times more than the simulator. On a simulated routing task planned at a fixed 30\,Hz
control rate, this throughput converts directly into outcome: 98\%  against at most 30\% for the learned baselines.

Several directions follow from these results. Tasks where long-term
behavior matters, such as knot tying and untangling, are interesting applications for model-based control of DLOs. Studying embedding strategies to allow a pretrained model to generalize to ropes of arbitrary length is another direction for future work.
Finally, except for the embeddings $e^{(i)}$ used to identify the rope segments,
little in the formulation is specific to ropes: the state is a set of
points, the decoder is shared across them, and the action enters as a
local displacement, so the same construction may transfer to other
deformable objects.

\section*{Acknowledgements}

This work was partially funded by the Brazilian Ministry of Science, Technology, and Innovations, with resources from Law No. 8.248, of October 23, 1991, under the PPI-SOFTEX program, DOU 01245.003479/2024-10, H.IAAC.
We thank UNICAMP's AdRoLab under Prof. Eric Rohmer for providing the robots used for data collection and evaluation.

\bibliographystyle{ieeetr}
\bibliography{references}

\end{document}